\documentclass[sigconf, nonacm, urlbreakonhyphens=false]{acmart}

\AtBeginDocument{%
  }

\usepackage{hyperref}

\hypersetup{
    colorlinks=true,   
    linkcolor=blue,    
    urlcolor=blue,     
    citecolor=cyan     
}
\definecolor{citeblue}{RGB}{0, 0, 128} 
\usepackage[table]{xcolor}
\usepackage{colortbl}
\begin{document}

\title{Beyond Monologue: Interactive Talking-Listening Avatar Generation with Conversational Audio Context-Aware Kernels}


\author{Yuzhe Weng$^{1 \ast}$, Haotian Wang$^{1 \ast}$, Xinyi Yu$^1$, Xiaoyan Wu$^{2}$, \\ \vspace{0em}
Haoran Xu$^2$, Shan He$^2$, Jun Du$^{1 \dagger}$ \\ \vspace{0.2em}
$^1$ University of Science and Technology of China,  $^2$ iFLYTEK  \\
}

\begin{abstract}
Audio-driven human video generation has achieved remarkable success in monologue scenarios, largely driven by advancements in powerful video generation foundation models. Moving beyond monologues, authentic human communication is inherently a full-duplex interactive process, requiring virtual agents not only to articulate their own speech but also to react naturally to incoming conversational audio. Most existing methods simply extend conventional audio-driven paradigms to listening scenarios. However, relying on strict frame-to-frame alignment renders the model's response to long-range conversational dynamics rigid, whereas directly introducing global attention catastrophically degrades lip synchronization. Recognizing the unique temporal Scale Discrepancy between talking and listening behaviors, we introduce a multi-head Gaussian kernel to explicitly inject this physical intuition into the model as a progressive temporal inductive bias. Building upon this, we construct a full-duplex interactive virtual agent capable of simultaneously processing dual-stream audio inputs for both talking and listening. Furthermore, we introduce a rigorously cleaned Talking-Listening dataset VoxHear featuring perfectly decoupled speech and background audio tracks. Extensive experiments demonstrate that our approach successfully fuses strong temporal alignment with deep contextual semantics, setting a new state-of-the-art for generating highly natural and responsive full-duplex interactive digital humans. The project page is available at \href{https://warmcongee.github.io/beyond-monologue/}{\color{blue}{BeyondMonologue-Page}}
\end{abstract}

\maketitle
\begingroup
\renewcommand\thefootnote{} 
\footnotetext{
    \leftskip=-1.2em 
    $^*$Equal contribution. $^\dagger$Corresponding author: \texttt{jundu@ustc.edu.cn}. \\
}
\addtocounter{footnote}{-1} 
\endgroup
\pagestyle{plain} 

\section{Introduction}
Audio-driven human video generation has advanced significantly for applications like digital avatars, achieving state-of-the-art success primarily in "monologue" scenarios by generating highly realistic expressions and precise lip-sync from a single speech input. However, authentic human communication is inherently a bidirectional process. An interactive digital agent cannot merely act as an isolated speaker; it must simultaneously articulate its own speech while dynamically responding to its interlocutor’s audio. Therefore, extending video generation from isolated monologues to handling full-duplex "Talking-Listening Interaction" is a crucial step toward the next generation of interactive digital avatars.

Recent attempts\cite{tran2024dim} to build talking-listening interactive agents often formulate the task as a hard switch between speaking and listening, rendering them incapable of handling audio overlap in real-world scenarios. Other works\cite{zhu2025infp} have explored full-duplex generation strictly on talking heads, failing to generate upper-body or full-body reactive postures. Furthermore, recent interactive agents\cite{sun2025streamavatar} adopt the monologue paradigm by directly applying frame-level local attention to incoming listening audio, but this naive approach leads to rigid and unnatural interactions.

The fundamental reason lies in the distinct driving mechanisms behind talking and listening. Articulating speech demands instantaneous acoustic-visual mapping to ensure precise lip synchronization. Conversely, listening reactions are highly dependent on broader conversational semantics, prosody, and turn-taking cues over a longer temporal span. This creates a fundamental trade-off between high-precision temporal synchronization and long-range contextual understanding. Introducing global video-to-audio attention for conversational context significantly degrades lip-sync quality. Conversely, strictly employing frame-to-frame local attention to preserve lip alignment weakens temporal context, leading to lifeless and stereotyped interactive behaviors. Prior work (e.g., OVIS\cite{low2025ovi}) shows that global cross-modal attention is essential for coherent generation. However, in unidirectional audio-driven video generation, strict temporal alignment must also be preserved. This necessitates jointly modeling strong local alignment and global context, rather than treating them as mutually exclusive. Beyond these algorithmic challenges, the field also faces a severe data bottleneck: existing datasets often contain entangled, overlapping, or noisy audio from both speakers, limiting effective data-driven training.

To address these challenges, we observe that interactive audio in full-duplex human video generation exhibits a Dual-Resolution Property: it simultaneously requires fine-grained, hard temporal alignment for speech articulation and coarse-grained, soft contextual understanding for natural interactive behaviors. This suggests that talking–listening generation cannot rely solely on either strict local alignment or global contextual modeling, but instead requires both within a unified architecture.
Motivated by this insight, we design a human video generation framework that synchronously supports dual-stream control from speaking and listening audio, and train it in two stages: speaking first, followed by concurrent speaking and listening. To reconcile local precision with long-range context, we introduce a progressive temporal inductive bias through a multi-head Gaussian kernel, which distributes the receptive fields of attention heads from narrow to wide. Heads with narrow receptive fields focus on strongly correlated frame-level audio cues, supporting precise lip synchronization and rhythmic motion, while heads with wider receptive fields capture longer audio context to drive natural interaction-related behaviors. In this way, the model jointly learns strong local temporality and global contextual awareness within a single framework.

\begin{figure}[htbp] 
    \centering
    \includegraphics[width=\columnwidth]{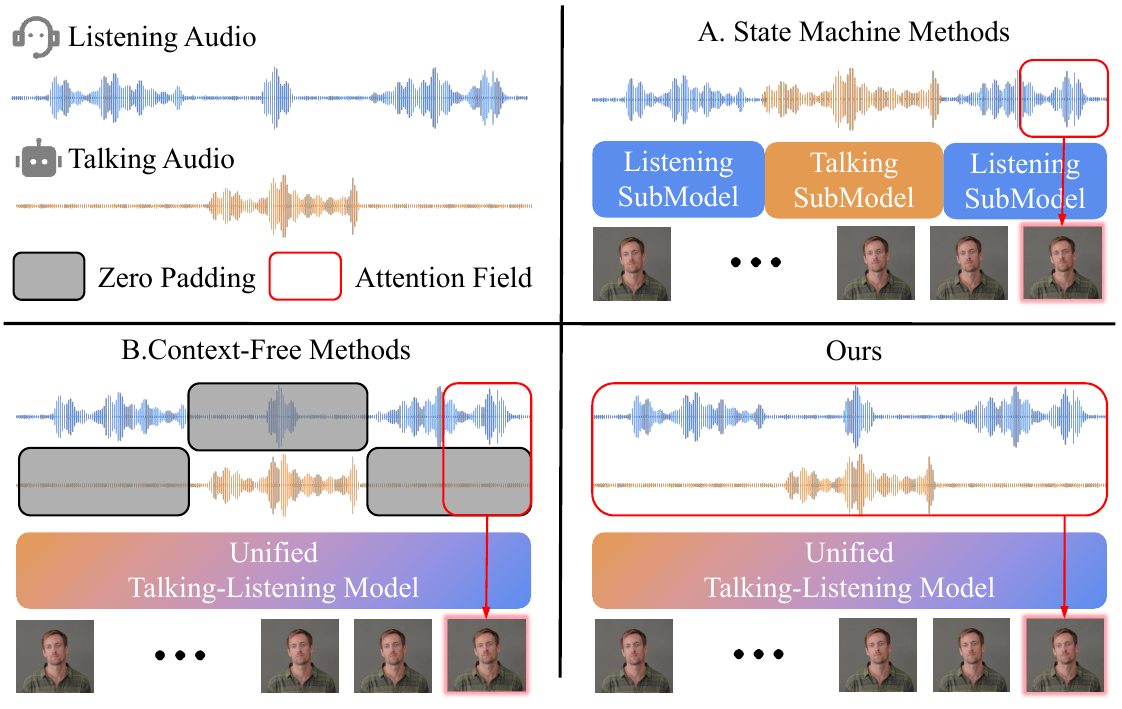} 
    \caption{Overview of our approach versus existing methods. Unlike state machine models (A, top right) that require manual switching, or context-free methods (B, bottom left) that rely on zero-padding and local attention, our unified model (bottom right) leverages global audio context awareness to achieve seamless, full-duplex interaction.}
    \label{fig:single_column_img}
    \vspace{-15pt}
\end{figure}

In summary, our main contributions are as follows:


We propose a unified attention architecture based on a multi-head Gaussian kernel. By enforcing differentiated temporal constraints, we resolve the inherent trade-off between precise lip synchronization and long-range audio context.

Our efficient talking–listening dual-stream human-centric video generation framework leverages arbitrary-position guidance and multi-scale audio context to achieve state-of-the-art performance in generating natural, full-duplex interactive avatars.

We construct a high-quality, large-scale interactive video dataset \textit{VoxHear} comprising 1{,}206 hours of cleanly decoupled talking and listening audio tracks, providing a robust foundation for data-driven, human-centric interactive video generation.


\section{Related Works}
\subsection{Audio-driven Human Video Generation}
Frontier technologies in video generation have seen explosive growth in recent years. As a downstream task, Audio-driven Human Video Generation has also consequently experienced a parallel boom\cite{cheng2024dawn, EmotiveTalk, tan2024flowvqtalker, jang2024faces, peng2024synctalk}. Early approaches\cite{StyleTalk, ma2023dreamtalk} often relied on predicting intermediate representations such as facial landmarks or 3D Morphable Models (3DMMs), which were then rendered into video. These works primarily focused on achieving precise lip synchronization with the input audio. Subsequent research further enhanced generative performance, producing more natural facial expressions and head movements.

More recently, diffusion-based approaches~\cite{tian2024emo,xu2024hallo,chen2025echomimic} built upon pre-trained image diffusion models for high-fidelity talking portrait synthesis, while newer methods such as EchoMimicV3~\cite{meng2026echomimicv3} further exploit DiT-based video diffusion models to improve realism.
While these methods have attained remarkable results in driving videos with monologue audio, they cannot react to listened audio or fully exploit audio context for human video generation.

\begin{figure*}[htbp]
    \centering
    \includegraphics[width=\textwidth]{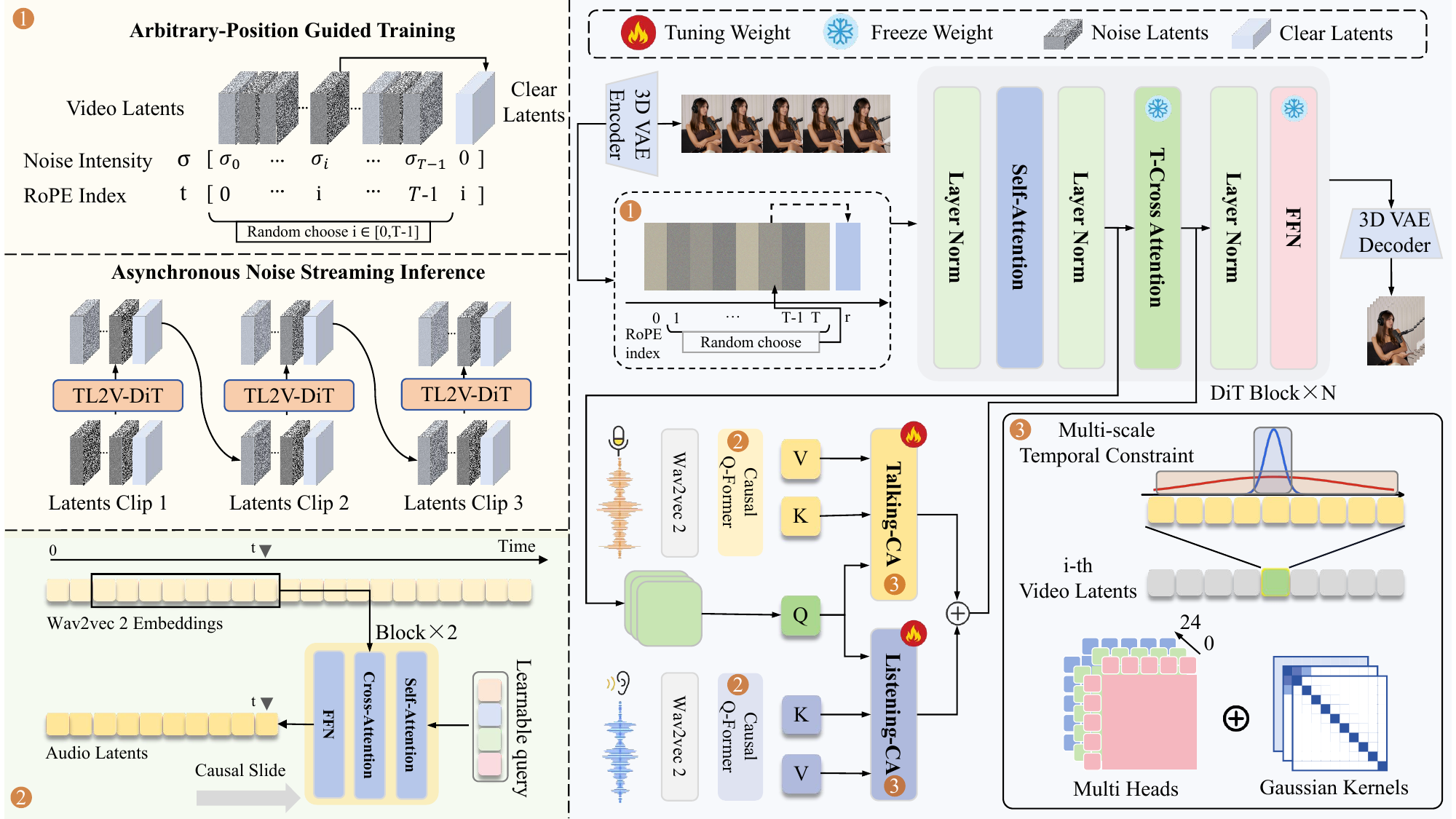} 
    \caption{Overview of the proposed framework. Top left: our training and inference scheme that unifies arbitrary-position reference guidance and diffusion forcing. Bottom right: the proposed causal Q-Former. Right: the architecture of the Talking-Listening dual-stream driving model with multi-scale Gaussian-constrained cross-modal interactions.}
    \label{fig:wide_image}
\end{figure*}

\subsection{Talking-Listening Video Generation}
Several works\cite{ng2022learning, liu2023mfr, ng2023can, zhou2022responsive,song2023emotional, song2024react, liu2024customlistener, tran2024dim, zhu2025infp, sun2025streamavatar} have recognized the crucial role of listened audio information in driving the natural responses of virtual humans, making extensive attempts in this direction. Constrained by small-scale networks and datasets\cite{zhou2022responsive, geng2023realtalk, luo2025omniresponse}, early two-stage methods\cite{liu2024customlistener, tran2024dim} achieved preliminary success by using generative models to translate audio and visual inputs into intermediate motions for rendering basic reactive portrait animations. To overcome the intermediate bottlenecks in two-stage generation that restrict motion expressivity and visual quality, exploring end-to-end generation\cite{sun2025streamavatar, luo2025omniresponse} is essential for achieving higher-quality generated videos.

Building on earlier studies of independent talking and listening, some prior works\cite{ng2022learning, ng2023can} model these two behaviors as mutually exclusive states, forcing virtual agents to switch strictly between speaking and listening and thus failing to handle natural audio overlap. More recent methods have attempted simultaneous talking and listening. For example, INFP\cite{zhu2025infp} leverages a codebook of intermediate motions to generate reasonable head poses, facial expressions, and lip movements, but remains limited to the head region and lacks broader expressiveness. StreamAvatar\cite{sun2025streamavatar} enables end-to-end talking-listening generation, yet simply applies the same local frame-to-frame 2D attention to listening audio as to talking audio, which severely limits its ability to capture conversational dynamics. In addition, its masking strategy for audio control during training introduces a noticeable train–test gap at inference time. In contrast, our approach emphasizes the significance of contextual information during the interactive talking-listening generation process. By balancing local frame-level audio-visual synchronization with global audio information, we achieve a stable, full-duplex talking-listening model with consistent training and inference paradigms.

\subsection{Interactive Audiovisual Datasets}
The advancement of data-driven interactive models depends heavily on the quality of the underlying datasets. Early works collected small-scale, head-only listening conversation datasets, including ViCo, RealTalk, and ResponseNet. These datasets feature limited identity diversity, restricted framing (head-only), and a general duration of under 10 hours, significantly constraining data-driven model training. Recent datasets, such as Seamless and SpeakerVid-5M, have collected larger-scale interactive talking-listening conversational data but are hampered by the entanglement and overlapping of talking and listening speech. To overcome this critical bottleneck, we designed a rigorous data construction and cleaning pipeline.

\section{Methods}

\subsection{Preliminary}
\label{sec:task_definition}

\textbf{Task Definition.} Let the ground-truth interactive human video be denoted as $\mathbf{X}_{1:F}$, where $F$ is the total number of frames. The full-duplex interactive video generation task takes a reference portrait image $I_{ref}$, a talking audio sequence $\mathbf{A}^{talk}_{1:F_a}$, and a listening audio sequence $\mathbf{A}^{listen}_{1:F_a}$ as inputs, and aims to generate a video sequence $\hat{\mathbf{X}}_{1:F}$. The goal is to synthesize a video in which the target identity accurately articulates the talking speech while simultaneously producing appropriate reactions to the listening audio.

\textbf{Flow Matching.} We adopt Flow Matching (FM)~\cite{lipman2022flow} with the Optimal Transport formulation. Given noise $x_0 \sim p_0(x)$ and data $x_1 \sim q(x_1)$, the interpolation path is
$$
x_t = (1 - t)x_0 + tx_1,
$$
with target velocity
$$
u_t(x_t) = x_1 - x_0.
$$
We use a Diffusion Transformer (DiT)~\cite{peebles2023dit}, parameterized by $\theta$, to predict the velocity field conditioned on $x_t$, $t$, and $c$, where $c$ includes the reference image $I_{ref}$ and dual-stream audio inputs. The training objective is
$$
\mathcal{L}_{FM} = \mathbb{E}_{t,x_0,x_1}\left[\|v_\theta(x_t,t,c)-(x_1-x_0)\|^2\right].
$$
During inference, starting from $x_0 \sim \mathcal{N}(0,I)$, we solve the ODE
$$
\frac{dx_t}{dt}=v_\theta(x_t,t,c)
$$
from $t=0$ to $t=1$ to obtain the generated video representation.

\subsection{Dual-Stream Architecture for Full-Duplex Generation}

In this section, we detail our Talking-Listening dual-stream audio-driven pipeline, built upon the mid-sized Wan 2.2 5B video generation model backbone\cite{wan2025wan}. Fundamentally, this is a Double A2V (Audio-to-Video) model built upon a DiT (Diffusion Transformer) architecture. It leverages the Wan2.2 VAE to achieve high video compression by a factor of 4×16×16. Furthermore, a two-stage incremental training scheme enables the model to switch seamlessly between a single-stream audio-driven mode and a dual-stream talking-listening-driven mode.

\subsubsection{\textbf{Adaptive Talking-Listening Audio Injection}}
Past work shows that different layers of pre-trained audio encoders capture different levels of information, with lower layers focusing more on phonetic details and higher layers encoding richer semantics. In dual-stream audio-driven video generation, the Talking and Listening branches rely on different audio cues, making a shared fixed-layer representation suboptimal. In addition, existing Classifier-Free Guidance (CFG) methods often use all-zero vectors as unconditional audio inputs, which introduces an out-of-distribution signal and increases the risk of visual artifacts under large guidance scales.

To address this, we design independent Audio Q-Formers for the Talking and Listening streams as learnable modules for feature compression and cross-modal alignment. Given an input audio, a pre-trained Wav2Vec 2.0 encoder produces representations from all layers, which we concatenate and project to a compact dimension $D_a$. We then divide the sequence into $W$ overlapping temporal windows aligned with the video latents, and use $N$ learnable queries within each window to aggregate audio features through cross-attention. The two branches use independent Q-Former weights, allowing them to attend to different hierarchical audio cues according to their respective roles.

This design provides two key advantages. First, it enables \textbf{adaptive fusion of multi-layer representations}, allowing the Talking and Listening branches to automatically discover task-specific cross-layer feature combinations. Second, it produces \textbf{smoothed unconditional embeddings for improved CFG}~\cite{ho2022cfg}. Instead of relying on artificial all-zero inputs, the Q-Former generates unconditional embeddings within its learned space, substantially narrowing the gap to conditional representations and allowing higher guidance scales without introducing visual artifacts such as color aberrations.

We freeze the video-text cross-attention. For each layer, following the IP-Adapter paradigm, we inject information into the video from both the Talking and Listening audios via 3D \textbf{Spatiotemporal Cross-Attention}. The Talking and Listening audio representations employ 1D RoPE encodings with identical indices at corresponding positions and share the same video Query, thereby maintaining temporal information and architectural consistency.

\begin{figure}[t] 
    \centering
    \includegraphics[width=\columnwidth]{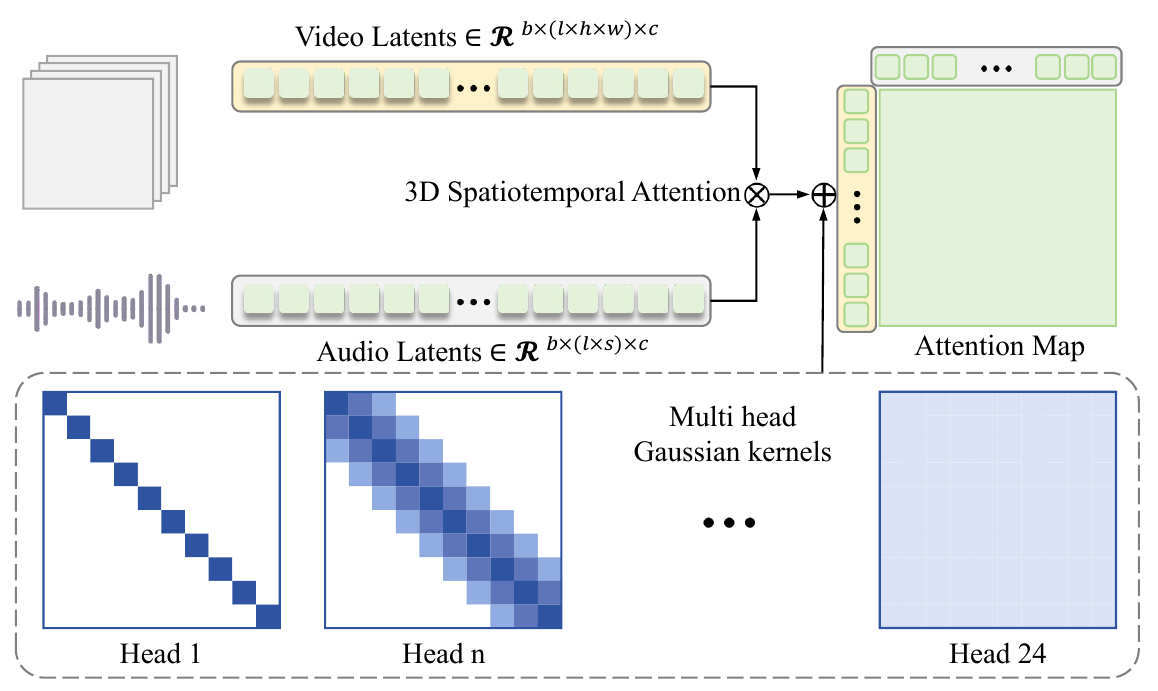} 
    \caption{Audio-Visual 3D Spatiotemporal Cross-Attention with Multi-Head Gaussian Kernels for Multi-Scale Temporal Modeling}
    \vspace{-10px}
    \label{fig:single_column_img}
\end{figure}

\subsubsection{\textbf{Temporal Constraint via Multi-Scale Gaussian Kernel}}

In audio-driven video generation and Talking-Listening dual-stream interactive tasks, cross-modal attention mechanisms face the classic local-global trade-off: 2D spatial cross-attention forces audio-visual alignment within a single frame, which, while achieving precise lip synchronization, completely severs the global context. Conversely, although 3D spatiotemporal cross-attention preserves the global receptive field, its fully connected attention matrix is excessively sparse. It neglects the strong temporal physical alignment prior inherent between audio and video, inevitably leading to an objective degradation in lip alignment accuracy.

To overcome this dilemma, we propose a Multi-Head Gaussian Kernels (MHGK) architecture that introduces multi-scale temporal constraints. First, we define the video latent sequence as $X \in \mathbb{R}^{B \times (L \times H \cdot W) \times D}$, and the corresponding audio temporal representation as $C \in \mathbb{R}^{B \times (L_a \times S) \times D}$. Here, $B$ denotes the batch size, $L$ and $L_a$ denote the temporal lengths of the video and audio respectively, $H \cdot W$ is the number of spatial latents per video frame, $S$ is the number of audio latents per frame, and $D$ represents the feature dimension. We adopt 3D spatiotemporal attention as our foundational framework. To address the sequence length mismatch between $L$ and $L_a$, we apply a temporally scaled, consistent 1D Rotary Position Embedding (1D RoPE) to both modalities. This scales them to the same temporal unit index $t$, endowing them with preliminary temporal alignment capabilities.

\begin{figure*}[t] 
    \centering
    \includegraphics[width=0.8\textwidth]{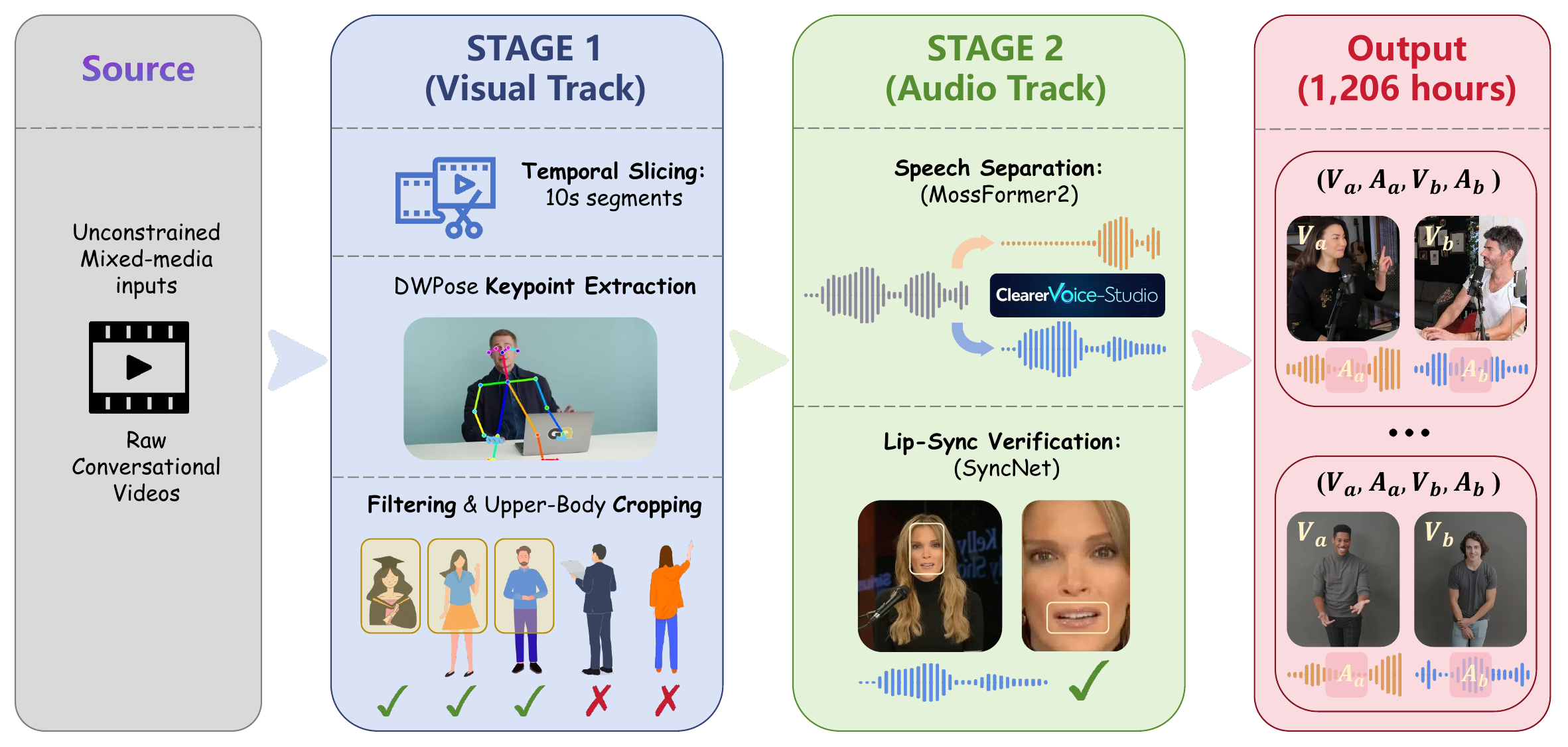} 
    \caption{Our cleaning and filtering pipeline for the VoxHear dataset consists of two stages: Visual Track filtering and Audio Track filtering.}
    \label{fig:my_wide_image}
\end{figure*}


To resolve the sparsity issue induced by global attention, we introduce the Multi-Head Gaussian Kernels into the attention computation. Specifically, when calculating cross-modal attention, we introduce an explicit Gaussian inductive bias matrix $\mathcal{B}^{(h)}$ as a distance penalty term for each attention head $h$:
$$
\text{Attention}^{(h)}(Q, K, V) = \text{Softmax}\left(\frac{Q^{(h)}(K^{(h)})^T}{\sqrt{d}} - \mathcal{B}^{(h)}\right)V^{(h)}
$$
where $\mathcal{B}^{(h)}$ is the Gaussian inductive bias matrix for the $h$-th attention head. To ensure maximum alignment for features with proximate temporal indices, the bias term is designed as a Gaussian distance penalty function that decays as temporal distance increases:
$$
\mathcal{B}^{(h)}(i, j) = \alpha_h \left( 1 - \exp\left(-\frac{(i - j)^2}{2\sigma_h^2}\right) \right)
$$
where $i$ and $j$ represent the aligned temporal indices of the video and audio, respectively. By assigning different standard deviations $\sigma_h$ and scaling coefficients $\alpha_h$ to different attention heads, the model achieves dynamic allocation of receptive fields. For attention heads with an extremely small standard deviation ($\sigma_h \to 0$), the Gaussian distribution becomes highly steep and $\alpha_h$ is maximized; the mechanism degrades into local attention with an extremely narrow receptive field, forcing the model to learn strict lip alignment. Conversely, for attention heads with an extremely large standard deviation ($\sigma_h \to \infty$), the penalty term $\mathcal{B}^{(h)} \to 0$, thereby losslessly preserving the 3D global attention to capture the contextual emotion and semantics of the audio.

Our method offers two key advantages. First, MHGK introduces an elegant temporal inductive bias that injects the physical prior of audio-visual alignment while enabling multi-scale modeling across attention heads, from narrow local lip-shape cues to broad global context. Second, it incurs negligible computational overhead, since the Gaussian bias operates directly on the attention score matrix without requiring extra modules such as temporal convolutions or additional local cross-attention layers. Compared with 2D Spatial CA, 3D CA + 1D RoPE, and 3D CA + 1D RoPE + ALiBi\cite{press2021train}, our 3D CA + 1D RoPE + MHGK achieves a better balance between lip synchronization accuracy and audio-visual prosody.

\subsection{Arbitrary-Position Guided Training and Inference Strategy}

Early methods mainly rely on fixed-position anchoring for scene and identity control. First-frame anchoring~\cite{wang2026read,wang2025fantasytalking} often suffers from the Attention Sink effect, causing later frames to over-attend to the first frame, which restricts motion diversity and increases the risk of identity drift in long-range generation. Although later works adopt last-frame anchoring to guide generation toward the temporal end, both strategies still depend on a single fixed temporal pattern, limiting global dynamic modeling.

To overcome this limitation, we adopt an arbitrary-position conditional guidance paradigm that balances generation stability and dynamic expressiveness. During training, for a noisy video clip of length $T$, we randomly sample an arbitrary frame $i \in [1, T]$, extract its clean latent $z_i$ as the guidance condition, and assign it the corresponding RoPE index $p_{ref}=i$. By removing the rigid constraint that guidance must come from the first or last frame, the model learns to generate coherent motion under more flexible temporal conditioning.

To further accommodate asynchronous noisy streaming during inference, we incorporate Diffusion Forcing into training by injecting different levels of random noise into different temporal chunks of the video latents. Combined with the clean guidance frame $z_i$ at an arbitrary position, this effectively turns the model into an \textbf{Arbitrary-position Video Expansion Model}. Rather than memorizing a fixed shortcut such as "diverging from head to tail" or "converging from tail to head," the model learns to perform temporally plausible expansion forward, backward, or bidirectionally from any absolute position on the timeline.

Full-duplex audio driving requires not only precise lip synchronization for Talking, but also natural listening responses for Listening. To stabilize the convergence of the dual-stream architecture and avoid early signal interference, we further adopt an incremental fine-tuning strategy. In \textbf{Stage 1 (Talking Priority)}, we introduce only the Talking Audio Adapter, allowing the model to first acquire accurate lip alignment and natural speaking behaviors. In \textbf{Stage 2 (Listening Fusion)}, we add the Listening Audio Adapter for further training, enabling the model to incorporate listening-aware interactive behaviors without disrupting the established talking capability.

An additional advantage of this training scheme is that it allows us to systematically explore the relationship between the guidance frame and the generation window during inference. We observe a clear trade-off: if the guidance frame is placed too far from the generation window, the model tends to weaken its reliance on the guidance signal, reducing identity retention; if it is placed too close, the generated content becomes less dynamically extensible. Because our training strategy exposes the model to arbitrary-position guidance, we can flexibly test and adopt different guidance distances at inference time. In particular, placing the reference image at a future position relative to the current generation window helps better balance identity preservation, motion amplitude, and natural character dynamics in long-term generation.

\begin{table*}[t]
\centering

\resizebox{\textwidth}{!}{
\definecolor{lightblue}{RGB}{218,232,252}
\begin{tabular}{l|ccc|c|ccc|ccc}
\toprule
 & \multicolumn{3}{c|}{\textbf{Perceptual Similarity}} & \textbf{Identity} & \multicolumn{3}{c|}{\textbf{Lip Synchronization}} & \multicolumn{3}{c}{\textbf{Video Quality}} \\
\textbf{Method} & \textbf{FID}$\downarrow$ & \textbf{FVD}$\downarrow$ & \textbf{LPIPS}$\downarrow$ & \textbf{CSIM}$\uparrow$ & \textbf{LMD}$\downarrow$ & \textbf{LSE-D}$\downarrow$ & \textbf{LSE-C}$\uparrow$ & \textbf{CPBD}$\uparrow$ & \textbf{ASE}$\uparrow$ & \textbf{IQA}$\uparrow$ \\
\midrule
GT              & 7.07 / 3.77 & 0.00 / 0.00 & 0.000 / 0.000 & 1.000 / 1.000 & 0.00 / 0.00 & 7.70 / 8.82 & 7.01 / 6.52 & 0.233 / 0.324 & 0.552 / 0.547 & 0.676 / 0.655 \\
\midrule
OmniAvatar\cite{gan2025omniavatar}      & \textbf{23.85} / 29.87 & \textbf{206.80} / 263.62 & 0.157 / 0.088 & 0.703 / 0.782 & 11.96 / 6.61 & \textbf{8.40} / 9.59 & \underline{6.50} / \underline{6.26} & 0.189 / 0.250 & 0.566 / \underline{0.549} & 0.666 / 0.617 \\
StableAvatar\cite{tu2025stableavatar}    & 25.92 / 91.61 & 269.76 / 623.22 & 0.171 / 0.206 & 0.681 / 0.659 & 13.30 / 9.97 & 11.72 / 13.09 & 2.68 / 2.26 & 0.197 / \textbf{0.361} & 0.556 / 0.487 & 0.662 / 0.558 \\
TalkVerse\cite{wang2025talkverse}       & 31.02 / 30.81 & 275.01 / 266.45 & 0.188 / 0.102 & 0.673 / 0.752 & 13.68 / 6.98 & 9.47 / 10.11 & 5.43 / 4.77 & \textbf{0.220} / \underline{0.282} & \textbf{0.574} / 0.548 & \textbf{0.678} / \textbf{0.637} \\
EchoMimic-v3\cite{meng2026echomimicv3}    & 25.92 / \underline{25.43} & 285.27 / \textbf{174.60} & 0.161 / \underline{0.071} & 0.687 / \underline{0.808} & 13.60 / 5.28 & 9.39 / \underline{9.51} & 5.27 / 5.69 & \underline{0.209} / 0.273 & 0.548 / 0.545 & \underline{0.675} / 0.624 \\
Fantasy-Talking\cite{wang2025fantasytalking} & 24.03 / 45.24 & 241.24 / 312.03 & \underline{0.149} / 0.108 & \underline{0.738} / 0.759 & \underline{11.73} / \underline{4.10} & 10.81 / 11.24 & 3.65 / 3.86 & 0.202 / 0.236 & 0.541 / 0.509 & 0.667 / 0.600 \\
Hallo3\cite{cui2025hallo3}          & 27.13 / 64.23 & 301.41 / 251.54 & 0.183 / 0.133 & 0.660 / 0.731 & 14.24 / 8.33 & 8.63 / 10.71 & 6.47 / 5.58 & 0.191 / 0.209 & 0.541 / 0.509 & 0.655 / 0.590 \\
\midrule
\rowcolor{lightblue} \textbf{Ours}   & \underline{23.96} / \textbf{21.82} & \underline{235.73} / \underline{206.33} & \textbf{0.145} / \textbf{0.057} & \textbf{0.749} / \textbf{0.876} & \textbf{10.25} / \textbf{3.48} & \underline{8.42} / \textbf{9.39} & \textbf{6.58} / \textbf{6.28} & 0.199 / 0.272 & \underline{0.573} / \textbf{0.556} & 0.666 / \underline{0.633} \\
\bottomrule
\end{tabular}
}
\caption{Quantitative comparison of various video generation models on the HDTF\cite{zhang2021flow} and MEAD\cite{kaisiyuan2020mead} datasets. Results are presented in the format of \textbf{HDTF / MEAD}. GT refers to the ground truth source video. $\uparrow$ indicates higher is better, and $\downarrow$ indicates lower is better. }
\label{tab:evaluation_results_combined}
\vspace{-20px}
\end{table*}

\subsection{High-Fidelity Decoupled Dataset Construction}

Training full-duplex speaking-listening video generation requires high-quality paired interaction data, where the audio of both participants is cleanly separated and temporally aligned with the visual content. However, existing conversational video datasets (e.g., Seamless\cite{agrawal2025seamless}, SpeakerVid-5M\cite{zhang2025speakervid}) often suffer from overlapping audio, multiple persons, severe occlusions, or full-body views, which weaken the learning signals for facial expressions and conversational dynamics.

To address these issues, we construct \textit{VoxHear} with a two-stage cleaning pipeline: (i) human-centric visual filtering and temporal standardization, and (ii) speech separation with lip-sync\cite{skorokhodov2022stylegan} validation. In the first stage, we remove clips with poor visual quality, inconsistent subject presence, or multiple detected persons, and further crop valid samples to focus on upper-body or portrait regions. In the second stage, we apply MossFormer2\cite{zhao2024mossformer2} from the ClearVoice\cite{zhao2025clearervoice} toolkit to separate mixed speech into independent single-speaker tracks, and then use SyncNet to verify that each separated audio track remains synchronized with the corresponding speaker's lip movements. Only samples that satisfy both visual and audio consistency checks are retained.

Following this pipeline, we obtain \textit{VoxHear}, a 1{,}206-hour speaking-listening interactive dataset. Each sample is represented as a temporally aligned quadruple $(V_a, A_a, V_b, A_b)$, where $V$ and $A$ denote the upper-body portrait video and independent audio track of each participant. Compared with existing resources, \textit{VoxHear} provides clean single-speaker audio, visually focused human-centered clips, and large-scale diverse interactive scenarios, offering a robust foundation for data-driven full-duplex human video generation.

\section{Experiments}

\subsection{Implementation Details}

Our model is trained in two stages, namely Talking training and Talking-Listening training, using different datasets in each stage. In the first stage, we train the Talking capability on several thousand hours of public and in-house collected data, with strict lip-sync alignment checking and DWPose-based human pose filtering for quality control. In the second stage, we further train the model on our self-constructed VoxHear dataset with over 1,000 hours of data.

Our model is built upon Wan2.2-5B and trained at 720p resolution with a multi-scale bucket-based dynamic-resolution strategy. We optimize the model using AdamW with bfloat16 mixed-precision training, and maintain an EMA model with a decay of 0.999. In both stages, we jointly optimize the newly introduced modules and the self-attention layers of the base model. The learning rate is set to 1e-5 for newly added parameters and 2e-6 for the trainable parameters in the backbone. All experiments are conducted on 16 NVIDIA A100 GPUs with a global batch size of 32. The first stage is trained for 100k steps, and the second stage for 30k steps.

\subsection{Overall Comparison}

\noindent\textbf{Baselines.}
We compare our method with six recent diffusion-based baselines, including OmniAvatar~\cite{gan2025omniavatar}, StableAvatar~\cite{tu2025stableavatar}, TalkVerse~\cite{wang2025talkverse}, EchoMimicV3~\cite{meng2026echomimicv3}, FantasyTalking~\cite{wang2025fantasytalking}, and Hallo3~\cite{cui2025hallo3}.
\noindent\textbf{Evaluation Metrics.} We evaluate our model using ten metrics from four aspects: perceptual similarity (FID~\cite{fid}, FVD~\cite{fvd}, LPIPS~\cite{LPIPS}), identity preservation (CSIM$\uparrow$~\cite{deng2019arcface}), lip synchronization (LMD$\downarrow$~\cite{chen2018lip}, LSE-D$\downarrow$, and LSE-C$\uparrow$ computed by SyncNet), and video quality (CPBD$\uparrow$~\cite{narvekar2011no}, ASE$\uparrow$~\cite{huang2023vbench}, and IQA$\uparrow$~\cite{huang2023vbench}).

\noindent\textbf{Quantitative Analysis.}
As shown in Table~\ref{tab:evaluation_results_combined}, our method achieves the best overall performance on both datasets, with clear advantages in identity preservation, lip synchronization, and perceptual similarity, while maintaining competitive video quality. In particular, StableAvatar and Fantasy-Talking show weak lip synchronization, TalkVerse favors video quality at the cost of identity consistency, and OmniAvatar, despite balanced performance, remains inferior to ours on the more challenging MEAD dataset.

As shown in Table~\ref{tab:taling_listening_metrics}, we reproduce DIM, the only publicly available modern baseline for Talking-Listening interactive generation, and our method consistently outperforms it across all metrics. Comparisons with other closed-source models are further supported by the user study presented later. Our model also generalizes well across diverse resolutions and portrait conditions. Furthermore, ablations on 3D spatiotemporal attention and the multi-scale Gaussian kernel validate the necessity of our design in the contextual setting of Talking-Listening interaction.


\begin{figure*}[t]
    \centering
    \includegraphics[width=\textwidth]{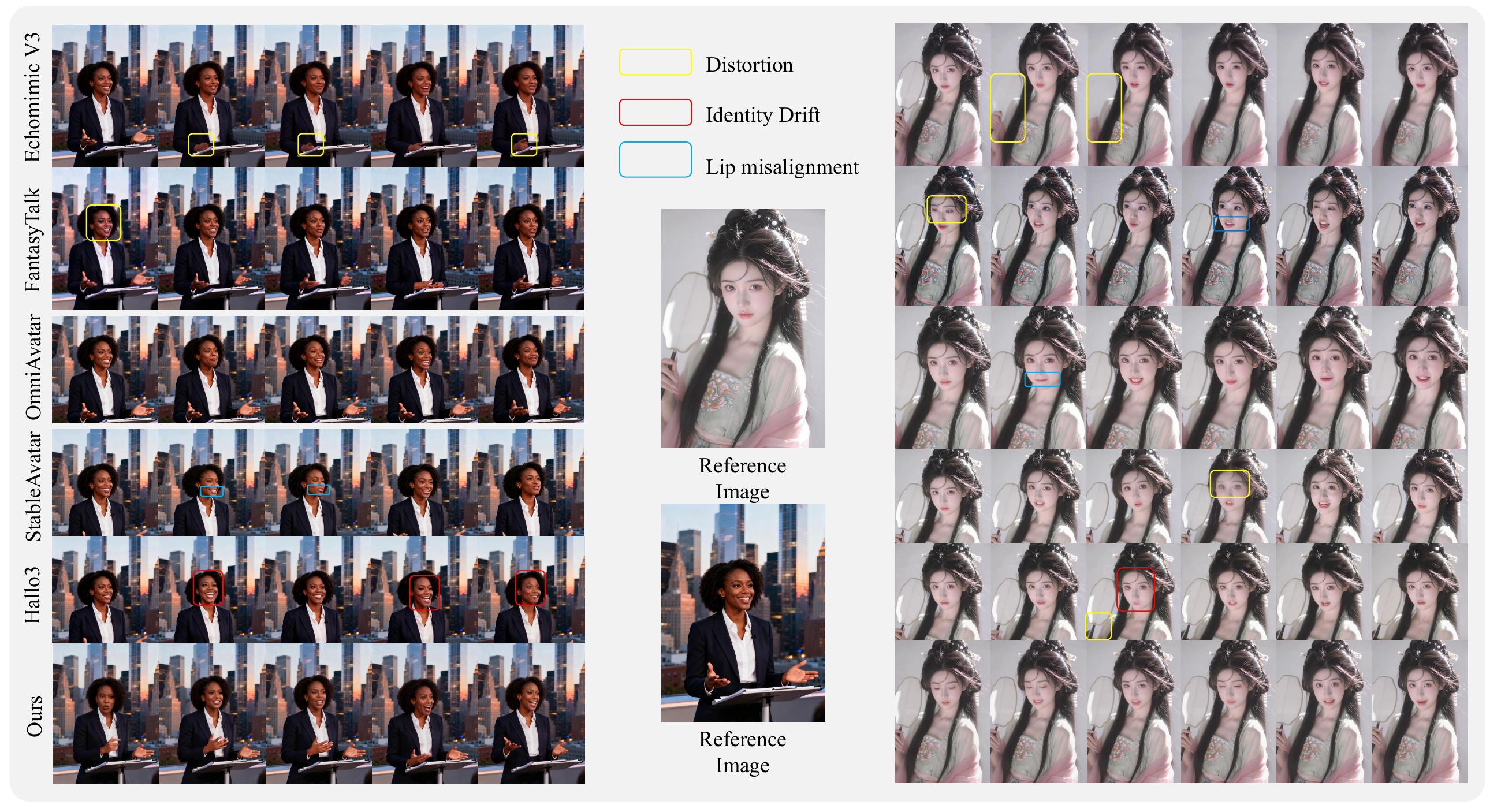} 
    \vspace{-15px}
    \caption{Qualitative comparison with state-of-the-art audio-driven video generation methods. The yellow boxes indicate visual distortions, the red boxes highlight identity drift, and the blue boxes mark lip misalignment.}
    
    \label{fig:overal_case_study}
\end{figure*}

\begin{table}[t]
\centering
\resizebox{\columnwidth}{!}{
\definecolor{lightblue}{RGB}{218,232,252}
\begin{tabular}{lccccc}
\toprule
\textbf{Method} & \textbf{CSIM}$\uparrow$ & \textbf{FID}$\downarrow$ & \textbf{FVD}$\downarrow$ & \textbf{LSE-C}$\uparrow$ & \textbf{ASE}$\uparrow$\\
\midrule
DIM\cite{tran2024dim} & 0.791 & 35.68 & 344.63 & 2.02 & 0.326 \\
\midrule
2D Spatial CA & 0.797 & 20.61 & 196.64 & 6.24 & 0.575\\
3D CA + 1D RoPE & 0.805 & 22.59 & 203.64 & 4.54 & 0.572\\
3D CA + 1D RoPE + ALiBi & 0.794 & 23.41 & 231.59 & 5.98 & \textbf{0.584}\\
\rowcolor{lightblue} Ours & \textbf{0.814} & \textbf{18.48} & \textbf{186.64} & \textbf{6.68} & 0.581\\
\bottomrule
\end{tabular}
}
\caption{Quantitative comparison and ablation studies for Talking-Listening interactive generation. We reproduce DIM, the only modern open-source method, and conduct all evaluations on the ResponseNet dataset.}
\vspace{-15pt}
\label{tab:taling_listening_metrics}
\end{table}

\subsection{Ablation Study}

\subsubsection{Attention Mechanisms}

We conduct an ablation study on the visual-audio attention fusion strategy. Specifically, we compare three baseline designs: \textit{2D Spatial CA}, \textit{3D CA + 1D RoPE}, and \textit{3D CA + 1D RoPE + ALiBi}, while keeping all other settings unchanged.

For \textit{2D Spatial CA}, following the common practice in previous audio-driven avatar video generation methods, we merge the temporal dimension $L$ into the batch dimension, so that each video latent only performs cross-attention with the audio latents within the corresponding current-frame temporal range for audio injection. For \textit{3D CA + 1D RoPE}, both video latents and audio latents are equipped with temporally scaled 1D RoPE before cross-attention, enabling temporal position modeling in the attention process. Based on this design, \textit{3D CA + 1D RoPE + ALiBi} further introduces a linear temporal decay bias to the attention scores of different heads, following prior work in the NLP literature. 
Our final method corresponds to the attention design described in Section~3. We note that the plain \textit{3D CA} variant without RoPE fails to learn meaningful audio-visual alignment and yields extremely poor results; therefore, we do not include it in Table \ref{tab:ablation_attention}.

\begin{table}[t]
\centering
\resizebox{\columnwidth}{!}{
\definecolor{lightblue}{RGB}{218,232,252}
\begin{tabular}{lcccc}
\toprule
\textbf{Method} & \textbf{CSIM}$\uparrow$ & \textbf{FID}$\downarrow$ & \textbf{FVD}$\downarrow$ & \textbf{LSE-C}$\uparrow$ \\
\midrule
2D Spatial CA              & 0.689 & 28.12 & 306.72 & 6.37 \\
3D CA + 1D RoPE            & 0.704 & 26.41 & 271.59 & 4.98 \\
3D CA + 1D RoPE + ALiBi    & 0.722 & 25.72 & 279.66 & 5.57 \\
\midrule
\rowcolor{lightblue} \textbf{Ours}              & \textbf{0.749} & \textbf{23.96} & \textbf{235.73} & \textbf{6.58} \\
\bottomrule
\end{tabular}
}
\caption{Ablation study on attention mechanisms. Results are presented in the format of \textbf{HDTF}.}
\vspace{-15pt}
\label{tab:ablation_attention}
\end{table}

As shown in Table~\ref{tab:ablation_attention}, our method achieves the best performance in both lip synchronization and identity preservation. Although \textit{2D Spatial CA} attains relatively competitive lip-sync performance, it produces notably worse FVD scores, indicating that it fails to adequately capture the dynamic prosodic alignment between the generated video and the driving audio. In addition, some results exhibit overly exaggerated lip movements, which also leads to degraded identity consistency. The \textit{3D CA + 1D RoPE} variant shows inferior lip-sync performance, suggesting that unrestricted attention over long audio contexts may dilute the supervision signal required for precise lip alignment. Adding ALiBi improves lip synchronization by imposing temporal constraints on attention, demonstrating the benefit of temporal locality. However, all metrics still remain inferior to our Gaussian-based design, highlighting the clear advantage of our method.



\begin{table}[t]
\centering
\resizebox{\columnwidth}{!}{
\definecolor{lightblue}{RGB}{218,232,252}
\begin{tabular}{lcccc}
\toprule
\textbf{Method} & \textbf{CSIM}$\uparrow$ & \textbf{FID}$\downarrow$ & \textbf{FVD}$\downarrow$ & \textbf{LSE-C}$\uparrow$ \\
\midrule
First Guide      & 0.614 & 32.84 & 347.65 & 6.01 \\
Index 21 Guide   & 0.736 & 24.48 & 267.82 & 6.24 \\
Index 27 Guide   & 0.711 & 28.17 & 316.62 & 6.08 \\
\rowcolor{lightblue} Index 22 Guide   & \textbf{0.749} & \textbf{23.96} & \textbf{235.73} & \textbf{6.58} \\
\bottomrule
\end{tabular}
}
\caption{Ablation study on guide index. Results are presented in the format of \textbf{HDTF}.}
\vspace{-18pt}
\label{tab:ablation_guide}
\end{table}

\subsubsection{Guiding Position Ablation}

During \textit{Arbitrary-Position Guided Training}, we enable dynamic adjustment of the guiding frame index at inference time by combining randomly positioned guiding frames with the diffusion forcing strategy during training. To study the effect of the guiding position, we conduct an ablation over different guide indices, as summarized in Table~\ref{tab:ablation_guide}.

As shown in the table, using the first frame as the guide leads to the worst overall performance, especially in identity consistency. We attribute this to the fact that the first-frame setting does not sufficiently reinforce the attention sink behavior during training, making the generation process more prone to content drift and instability. When the RoPE index is set to 21, the guiding frame is placed too close to the denoising window. This causes the generated video to over-converge toward the guiding frame, which restricts motion dynamics and consequently degrades both FVD and lip-sync performance. In contrast, when the guiding frame is placed too far away, such as with index 27, the generated video becomes less stable and exhibits oscillatory artifacts, indicating that the model cannot adequately absorb the guiding information. As a result, both FID and FVD deteriorate noticeably. Among all settings, \textit{Index 22 Guide} achieves the best overall performance across all metrics, yielding the strongest identity preservation, visual quality, temporal coherence, and lip synchronization. These results validate our choice of guiding position for inference.

\begin{figure}[t]
    \centering
    \includegraphics[width=\columnwidth]{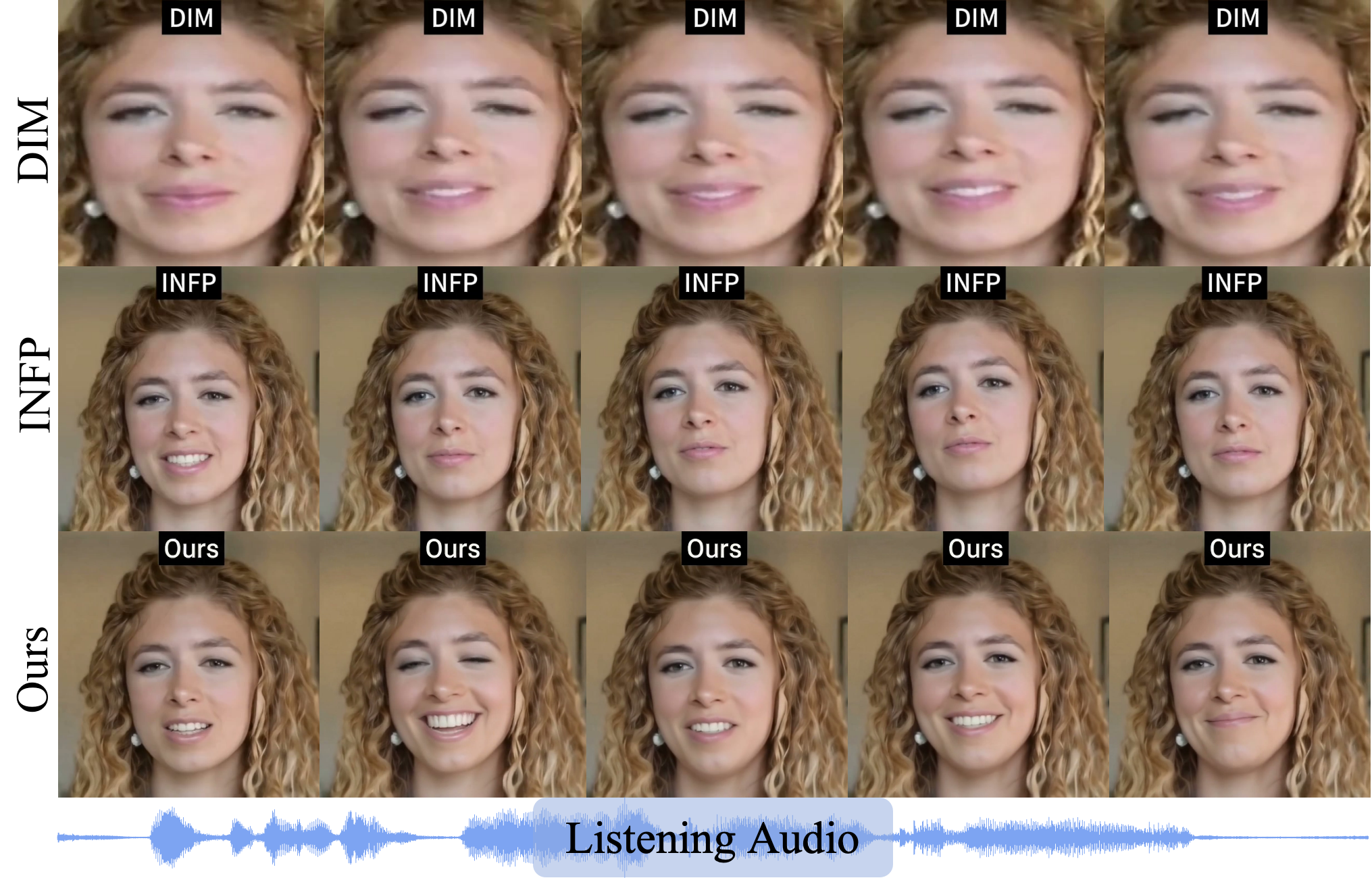}
    \caption{Qualitative comparison with other methods in the scenario where only the Listening Audio is present and the Talking Audio is nearly silent. } 
    \vspace{-5pt}
    \label{fig:listening_exp}
\end{figure}

\begin{table}[t]
\centering

\label{tab:batch4}
\resizebox{\columnwidth}{!}{
\definecolor{lightblue}{RGB}{218,232,252}
\begin{tabular}{lcccc}
\toprule
\textbf{Model} & \textbf{Natural.} $\uparrow$ & \textbf{Motion} $\uparrow$ & \textbf{AV Align.} $\uparrow$ & Visual $\uparrow$ \\
\midrule
RLHG\cite{zhou2022responsive}  & 1.41 & 1.36 & 1.68 &
1.50 \\
L2L\cite{ng2022learning}   & 1.59 & 1.45 & 1.77 &
1.73 \\
DIM\cite{tran2024dim}   & 1.68 & 2.05 & 2.00 &
1.86 \\
INFP\cite{zhu2025infp}  & 3.86 & 4.00 & 4.05 & \textbf{4.55} \\
\rowcolor{lightblue} Ours  & \textbf{4.14} & \textbf{4.05} & \textbf{4.18} & 4.32 \\
\bottomrule
\end{tabular}
}
\caption{User study results on listening.}
\vspace{-15pt}
\label{tab:user_study}
\end{table}

\begin{figure}[t]
    \centering
    \includegraphics[width=\columnwidth]{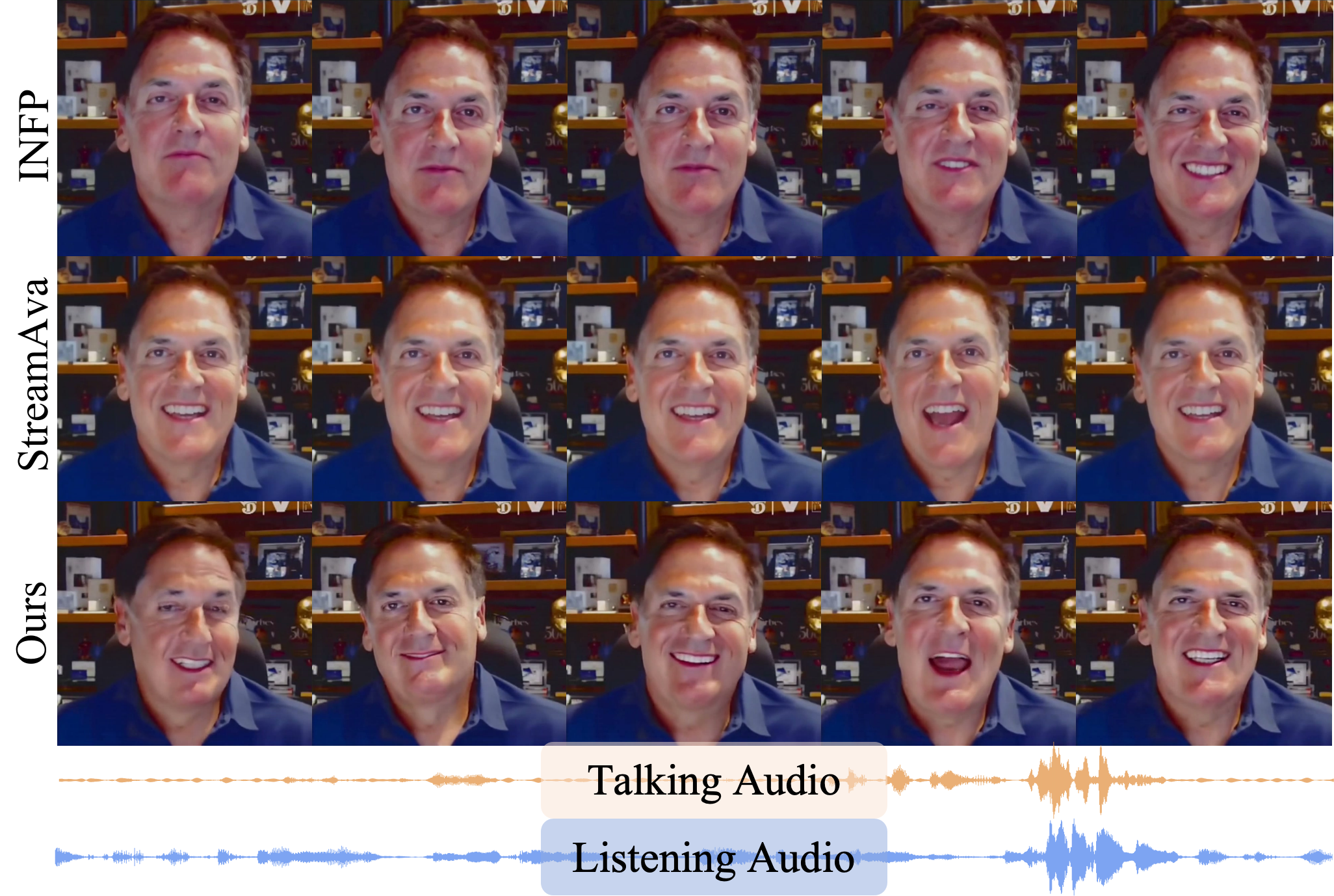}
    \caption{Qualitative comparison with other methods in full-duplex talking-listening interaction.}
    \vspace{-15pt}
    \label{fig:talking_listening_exp}
\end{figure}

\subsection{Case Study}
\textbf{Qualitative Analysis of Audio-Driven Video Generation.} 

As shown in Fig.~\ref{fig:overal_case_study}, we qualitatively compare our method with several state-of-the-art audio-driven human video generation models. EchoMimic exhibits noticeable distortions when handling challenging elements such as hands and fans. FantasyTalk suffers from occasional frame distortions and lip-sync issues. OmniAvatar achieves relatively good visual quality, but its body motions are limited and rigid. StableAvatar shows both visual distortions and lip-sync artifacts. Hallo3 has severe limitations in preserving character identity. Overall, our method delivers higher overall quality in pose control, lip articulation, and visual fidelity.

\textbf{Qualitative Analysis of Talking-Listening Interaction.} 
We compare our method with state-of-the-art models for talking-listening audio-driven generation. Since some SOTA methods are not open-sourced, we use their publicly available video demos for comparison. Figure~\ref{fig:listening_exp} evaluates the listening setting, where the Talking audio is silent and only the Listening audio is provided. Our method produces more natural and expressive facial responses and body dynamics than DIM and INFP within the same time interval, while also showing robust adaptability to driving images of different resolutions. Figure~\ref{fig:talking_listening_exp} evaluates the full-duplex talking-listening scenario. Compared with baselines, our method generates more natural and richer head motions and facial expressions, while INFP shows weaker alignment with the speaking content in the highlighted region.

\textbf{User Study}
We invited 11 participants to conduct a MOS evaluation of the results. Specifically, they rated each video on a scale from 1 to 5 across four aspects: naturalness, motion diversity, audio-visual lip synchronization, and visual quality. As shown, our method achieves the best performance on the first three metrics. In contrast, the compared models show relatively similar visual quality, making the differences difficult to distinguish by human perception.

\section{Conclusion and Future Work}
In this paper, we study full-duplex audio-driven human video generation, where a virtual agent must simultaneously speak and respond to conversational audio. We propose a unified talking-listening framework with a multi-head Gaussian kernel for balancing lip synchronization and long-range contextual modeling, and build a dual-stream architecture for natural interactive video generation. We also construct VoxHear, a large-scale carefully curated dataset with decoupled talking and listening audio tracks. Extensive experiments demonstrate strong realism, responsiveness, and temporal coherence.
More broadly, our work offers a new perspective on long-context audio understanding and alignment in video generation, enabling more vivid conversational behaviors. Nevertheless, the ultimate goal of this task is far more ambitious: a digital human should eventually be able to perceive spoken instructions and produce arbitrary human-like behaviors in an end-to-end manner. Reaching this vision will require substantial further exploration in both data and modeling.


\bibliographystyle{ACM-Reference-Format}
\bibliography{sample-base}

@String{Computer = "{IEEE} Computer" }

@String{Springer = "Springer-Verlag" }

@article{cheng2024dawn,
  title={DAWN: Dynamic Frame Avatar with Non-autoregressive Diffusion Framework for Talking Head Video Generation},
  author={Cheng, Hanbo and Lin, Limin and Liu, Chenyu and Xia, Pengcheng and Hu, Pengfei and Ma, Jiefeng and Du, Jun and Pan, Jia},
  journal={arXiv preprint arXiv:2410.13726},
  year={2024}
}

@inproceedings{EmotiveTalk,
  author       = {Haotian Wang and
                  Yuzhe Weng and
                  Yueyan Li and
                  Zilu Guo and
                  Jun Du and
                  Shutong Niu and
                  Jiefeng Ma and
                  Shan He and
                  Xiaoyan Wu and
                  Qiming Hu and
                  Bing Yin and
                  Cong Liu and
                  Qingfeng Liu},
  title        = {EmotiveTalk: Expressive Talking Head Generation through Audio Information
                  Decoupling and Emotional Video Diffusion},
  booktitle    = {{IEEE/CVF} Conference on Computer Vision and Pattern Recognition,
                  {CVPR} 2025, Nashville, TN, USA, June 11-15, 2025},
  pages        = {26212--26221},
  publisher    = {Computer Vision Foundation / {IEEE}},
  year         = {2025},
  url          = {https://openaccess.thecvf.com/content/CVPR2025/html/Wang\_EmotiveTalk\_Expressive\_Talking\_Head\_Generation\_through\_Audio\_Information\_Decoupling\_and\_CVPR\_2025\_paper.html},
  doi          = {10.1109/CVPR52734.2025.02441},
  bibsource    = {dblp computer science bibliography, https://dblp.org}
}

@inproceedings{tan2024flowvqtalker,
  title={Flowvqtalker: High-quality emotional talking face generation through normalizing flow and quantization},
  author={Tan, Shuai and Ji, Bin and Pan, Ye},
  booktitle={Proceedings of the IEEE/CVF Conference on Computer Vision and Pattern Recognition},
  pages={26317--26327},
  year={2024}
}

@inproceedings{jang2024faces,
  title={Faces that speak: Jointly synthesising talking face and speech from text},
  author={Jang, Youngjoon and Kim, Ji-Hoon and Ahn, Junseok and Kwak, Doyeop and Yang, Hong-Sun and Ju, Yoon-Cheol and Kim, Il-Hwan and Kim, Byeong-Yeol and Chung, Joon Son},
  booktitle={Proceedings of the IEEE/CVF Conference on Computer Vision and Pattern Recognition},
  pages={8818--8828},
  year={2024}
}

@inproceedings{peng2024synctalk,
  title={Synctalk: The devil is in the synchronization for talking head synthesis},
  author={Peng, Ziqiao and Hu, Wentao and Shi, Yue and Zhu, Xiangyu and Zhang, Xiaomei and Zhao, Hao and He, Jun and Liu, Hongyan and Fan, Zhaoxin},
  booktitle={Proceedings of the IEEE/CVF Conference on Computer Vision and Pattern Recognition},
  pages={666--676},
  year={2024}
}

@inproceedings{StyleTalk,
author = {Ma, Yifeng and Wang, Suzhen and Hu, Zhipeng and Fan, Changjie and Lv, Tangjie and Ding, Yu and Deng, Zhidong and Yu, Xin},
title = {StyleTalk: one-shot talking head generation with controllable speaking styles},
year = {2023},
isbn = {978-1-57735-880-0},
publisher = {AAAI Press},
url = {https://doi.org/10.1609/aaai.v37i2.25280},
doi = {10.1609/aaai.v37i2.25280},
booktitle = {Proceedings of the Thirty-Seventh AAAI Conference on Artificial Intelligence and Thirty-Fifth Conference on Innovative Applications of Artificial Intelligence and Thirteenth Symposium on Educational Advances in Artificial Intelligence},
articleno = {211},
numpages = {9},
series = {AAAI'23/IAAI'23/EAAI'23}
}

@article{ma2023dreamtalk,
  title={DreamTalk: When Expressive Talking Head Generation Meets Diffusion Probabilistic Models},
  author={Ma, Yifeng and Zhang, Shiwei and Wang, Jiayu and Wang, Xiang and Zhang, Yingya and Deng, Zhidong},
  journal={arXiv preprint arXiv:2312.09767},
  year={2023}
}

@inproceedings{tian2024emo,
  title={Emo: Emote portrait alive generating expressive portrait videos with audio2video diffusion model under weak conditions},
  author={Tian, Linrui and Wang, Qi and Zhang, Bang and Bo, Liefeng},
  booktitle={European Conference on Computer Vision},
  pages={244--260},
  year={2024},
  organization={Springer}
}

@article{xu2024hallo,
  title={Hallo: Hierarchical audio-driven visual synthesis for portrait image animation},
  author={Xu, Mingwang and Li, Hui and Su, Qingkun and Shang, Hanlin and Zhang, Liwei and Liu, Ce and Wang, Jingdong and Yao, Yao and Zhu, Siyu},
  journal={arXiv preprint arXiv:2406.08801},
  year={2024}
}

@inproceedings{chen2025echomimic,
  title={Echomimic: Lifelike audio-driven portrait animations through editable landmark conditions},
  author={Chen, Zhiyuan and Cao, Jiajiong and Chen, Zhiquan and Li, Yuming and Ma, Chenguang},
  booktitle={Proceedings of the AAAI Conference on Artificial Intelligence},
  volume={39},
  number={3},
  pages={2403--2410},
  year={2025}
}

@inproceedings{tran2024dim,
  title={Dim: Dyadic interaction modeling for social behavior generation},
  author={Tran, Minh and Chang, Di and Siniukov, Maksim and Soleymani, Mohammad},
  booktitle={European Conference on Computer Vision},
  pages={484--503},
  year={2024},
  organization={Springer}
}

@inproceedings{zhu2025infp,
  title={INFP: Audio-driven interactive head generation in dyadic conversations},
  author={Zhu, Yongming and Zhang, Longhao and Rong, Zhengkun and Hu, Tianshu and Liang, Shuang and Ge, Zhipeng},
  booktitle={Proceedings of the IEEE/CVF Conference on Computer Vision and Pattern Recognition},
  pages={10667--10677},
  year={2025}
}

@article{sun2025streamavatar,
  title={StreamAvatar: Streaming Diffusion Models for Real-Time Interactive Human Avatars},
  author={Sun, Zhiyao and Peng, Ziqiao and Ma, Yifeng and Chen, Yi and Zhou, Zhengguang and Zhou, Zixiang and Zhang, Guozhen and Zhang, Youliang and Zhou, Yuan and Lu, Qinglin and others},
  journal={arXiv preprint arXiv:2512.22065},
  year={2025}
}

@inproceedings{liu2023mfr,
  title={Mfr-net: Multi-faceted responsive listening head generation via denoising diffusion model},
  author={Liu, Jin and Wang, Xi and Fu, Xiaomeng and Chai, Yesheng and Yu, Cai and Dai, Jiao and Han, Jizhong},
  booktitle={Proceedings of the 31st ACM international conference on multimedia},
  pages={6734--6743},
  year={2023}
}

@inproceedings{liu2024customlistener,
  title={Customlistener: Text-guided responsive interaction for user-friendly listening head generation},
  author={Liu, Xi and Guo, Ying and Zhen, Cheng and Li, Tong and Ao, Yingying and Yan, Pengfei},
  booktitle={Proceedings of the IEEE/CVF Conference on Computer Vision and Pattern Recognition},
  pages={2415--2424},
  year={2024}
}

@inproceedings{ng2022learning,
  title={Learning to listen: Modeling non-deterministic dyadic facial motion},
  author={Ng, Evonne and Joo, Hanbyul and Hu, Liwen and Li, Hao and Darrell, Trevor and Kanazawa, Angjoo and Ginosar, Shiry},
  booktitle={Proceedings of the IEEE/CVF conference on computer vision and pattern recognition},
  pages={20395--20405},
  year={2022}
}

@inproceedings{ng2023can,
  title={Can language models learn to listen?},
  author={Ng, Evonne and Subramanian, Sanjay and Klein, Dan and Kanazawa, Angjoo and Darrell, Trevor and Ginosar, Shiry},
  booktitle={Proceedings of the IEEE/CVF International Conference on Computer Vision},
  pages={10083--10093},
  year={2023}
}

@inproceedings{song2023emotional,
  title={Emotional listener portrait: Realistic listener motion simulation in conversation},
  author={Song, Luchuan and Yin, Guojun and Jin, Zhenchao and Dong, Xiaoyi and Xu, Chenliang},
  booktitle={2023 IEEE/CVF International Conference on Computer Vision (ICCV)},
  pages={20782--20792},
  year={2023},
  organization={IEEE}
}

@inproceedings{song2024react,
  title={React 2024: the second multiple appropriate facial reaction generation challenge},
  author={Song, Siyang and Spitale, Micol and Luo, Cheng and Palmero, Cristina and Barquero, German and Zhu, Hengde and Escalera, Sergio and Valstar, Michel and Baur, Tobias and Ringeval, Fabien and others},
  booktitle={2024 IEEE 18th International Conference on Automatic Face and Gesture Recognition (FG)},
  pages={1--5},
  year={2024},
  organization={IEEE}
}

@inproceedings{zhou2022responsive,
  title={Responsive listening head generation: a benchmark dataset and baseline},
  author={Zhou, Mohan and Bai, Yalong and Zhang, Wei and Yao, Ting and Zhao, Tiejun and Mei, Tao},
  booktitle={European conference on computer vision},
  pages={124--142},
  year={2022},
  organization={Springer}
}

@article{geng2023realtalk,
  title={Affective faces for goal-driven dyadic communication},
  author={Geng, Scott and Teotia, Revant and Tendulkar, Purva and Menon, Sachit and Vondrick, Carl},
  journal={arXiv preprint arXiv:2301.10939},
  year={2023}
}

@article{luo2025omniresponse,
  title={OmniResponse: Online Multimodal Conversational Response Generation in Dyadic Interactions},
  author={Luo, Cheng and Wang, Jianghui and Li, Bing and Song, Siyang and Ghanem, Bernard},
  journal={arXiv preprint arXiv:2505.21724},
  year={2025}
}

@article{low2025ovi,
  title={Ovi: Twin backbone cross-modal fusion for audio-video generation},
  author={Low, Chetwin and Wang, Weimin and Katyal, Calder},
  journal={arXiv preprint arXiv:2510.01284},
  year={2025}
}

@article{wan2025wan,
  title={Wan: Open and advanced large-scale video generative models},
  author={Wan, Team and Wang, Ang and Ai, Baole and Wen, Bin and Mao, Chaojie and Xie, Chen-Wei and Chen, Di and Yu, Feiwu and Zhao, Haiming and Yang, Jianxiao and others},
  journal={arXiv preprint arXiv:2503.20314},
  year={2025}
}

@article{lipman2022flow,
  title={Flow matching for generative modeling},
  author={Lipman, Yaron and Chen, Ricky TQ and Ben-Hamu, Heli and Nickel, Maximilian and Le, Matt},
  journal={arXiv preprint arXiv:2210.02747},
  year={2022}
}

@inproceedings{peebles2023dit,
  title={Scalable diffusion models with transformers},
  author={Peebles, William and Xie, Saining},
  booktitle={Proceedings of the IEEE/CVF international conference on computer vision},
  pages={4195--4205},
  year={2023}
}

@article{ho2022cfg,
  title={Classifier-free diffusion guidance},
  author={Ho, Jonathan and Salimans, Tim},
  journal={arXiv preprint arXiv:2207.12598},
  year={2022}
}

@article{press2021train,
  title={Train short, test long: Attention with linear biases enables input length extrapolation},
  author={Press, Ofir and Smith, Noah A and Lewis, Mike},
  journal={arXiv preprint arXiv:2108.12409},
  year={2021}
}

@inproceedings{wang2026read,
  title={Read: Real-time and efficient asynchronous diffusion for audio-driven talking head generation},
  author={Wang, Haotian and Weng, Yuzhe and Du, Jun and Xu, Haoran and Wu, Xiaoyan and He, Shan and Yin, Bing and Liu, Cong and Gao, Jianqing and Liu, Qingfeng},
  booktitle={Proceedings of the AAAI Conference on Artificial Intelligence},
  volume={40},
  number={12},
  pages={9766--9774},
  year={2026}
}

@inproceedings{wang2025fantasytalking,
  title={Fantasytalking: Realistic talking portrait generation via coherent motion synthesis},
  author={Wang, Mengchao and Wang, Qiang and Jiang, Fan and Fan, Yaqi and Zhang, Yunpeng and Qi, Yonggang and Zhao, Kun and Xu, Mu},
  booktitle={Proceedings of the 33rd ACM International Conference on Multimedia},
  pages={9891--9900},
  year={2025}
}

@article{agrawal2025seamless,
  title={Seamless interaction: Dyadic audiovisual motion modeling and large-scale dataset},
  author={Agrawal, Vasu and Akinyemi, Akinniyi and Alvero, Kathryn and Behrooz, Morteza and Buffalini, Julia and Carlucci, Fabio Maria and Chen, Joy and Chen, Junming and Chen, Zhang and Cheng, Shiyang and others},
  journal={arXiv preprint arXiv:2506.22554},
  year={2025}
}

@article{zhang2025speakervid,
  title={Speakervid-5m: A large-scale high-quality dataset for audio-visual dyadic interactive human generation},
  author={Zhang, Youliang and Li, Zhaoyang and Wang, Duomin and Zhang, Jiahe and Zhou, Deyu and Yin, Zixin and Dai, Xili and Yu, Gang and Li, Xiu},
  journal={arXiv preprint arXiv:2507.09862},
  year={2025}
}

@inproceedings{zhao2024mossformer2,
  title={Mossformer2: Combining transformer and rnn-free recurrent network for enhanced time-domain monaural speech separation},
  author={Zhao, Shengkui and Ma, Yukun and Ni, Chongjia and Zhang, Chong and Wang, Hao and Nguyen, Trung Hieu and Zhou, Kun and Yip, Jia Qi and Ng, Dianwen and Ma, Bin},
  booktitle={ICASSP 2024-2024 IEEE International Conference on Acoustics, Speech and Signal Processing (ICASSP)},
  pages={10356--10360},
  year={2024},
  organization={IEEE}
}

@article{zhao2025clearervoice,
  title={Clearervoice-studio: Bridging advanced speech processing research and practical deployment},
  author={Zhao, Shengkui and Pan, Zexu and Ma, Bin},
  journal={arXiv preprint arXiv:2506.19398},
  year={2025}
}

@inproceedings{skorokhodov2022stylegan,
  title={Stylegan-v: A continuous video generator with the price, image quality and perks of stylegan2},
  author={Skorokhodov, Ivan and Tulyakov, Sergey and Elhoseiny, Mohamed},
  booktitle={Proceedings of the IEEE/CVF conference on computer vision and pattern recognition},
  pages={3626--3636},
  year={2022}
}

@article{gan2025omniavatar,
  title={Omniavatar: Efficient audio-driven avatar video generation with adaptive body animation},
  author={Gan, Qijun and Yang, Ruizi and Zhu, Jianke and Xue, Shaofei and Hoi, Steven},
  journal={arXiv preprint arXiv:2506.18866},
  year={2025}
}

@article{tu2025stableavatar,
  title={Stableavatar: Infinite-length audio-driven avatar video generation},
  author={Tu, Shuyuan and Pan, Yueming and Huang, Yinming and Han, Xintong and Xing, Zhen and Dai, Qi and Luo, Chong and Wu, Zuxuan and Jiang, Yu-Gang},
  journal={arXiv preprint arXiv:2508.08248},
  year={2025}
}

@article{wang2025talkverse,
  title={TalkVerse: Democratizing Minute-Long Audio-Driven Video Generation},
  author={Wang, Zhenzhi and Wang, Jian and Ma, Ke and Lin, Dahua and Zhou, Bing},
  journal={arXiv preprint arXiv:2512.14938},
  year={2025}
}

@inproceedings{meng2026echomimicv3,
  title={Echomimicv3: 1.3 b parameters are all you need for unified multi-modal and multi-task human animation},
  author={Meng, Rang and Wang, Yan and Wu, Weipeng and Zheng, Ruobing and Li, Yuming and Ma, Chenguang},
  booktitle={Proceedings of the AAAI Conference on Artificial Intelligence},
  volume={40},
  number={10},
  pages={8008--8015},
  year={2026}
}

@inproceedings{zhang2021flow,
  title={Flow-guided one-shot talking face generation with a high-resolution audio-visual dataset},
  author={Zhang, Zhimeng and Li, Lincheng and Ding, Yu and Fan, Changjie},
  booktitle={Proceedings of the IEEE/CVF conference on computer vision and pattern recognition},
  pages={3661--3670},
  year={2021}
}

@inproceedings{kaisiyuan2020mead,
 author = {Wang, Kaisiyuan and Wu, Qianyi and Song, Linsen and Yang, Zhuoqian and Wu, Wayne and Qian, Chen and He, Ran and Qiao, Yu and Loy, Chen Change},
 title = {MEAD: A Large-scale Audio-visual Dataset for Emotional Talking-face Generation},
 booktitle = {ECCV},
 month = Augest,
 year = {2020}
}

@inproceedings{cui2025hallo3,
  title={Hallo3: Highly dynamic and realistic portrait image animation with video diffusion transformer},
  author={Cui, Jiahao and Li, Hui and Zhan, Yun and Shang, Hanlin and Cheng, Kaihui and Ma, Yuqi and Mu, Shan and Zhou, Hang and Wang, Jingdong and Zhu, Siyu},
  booktitle={Proceedings of the Computer Vision and Pattern Recognition Conference},
  pages={21086--21095},
  year={2025}
}

@article{fid,
  title={Gans trained by a two time-scale update rule converge to a local nash equilibrium},
  author={Heusel, Martin and Ramsauer, Hubert and Unterthiner, Thomas and Nessler, Bernhard and Hochreiter, Sepp},
  journal={Advances in neural information processing systems},
  volume={30},
  year={2017}
}

@article{fvd,
  title={FVD: A new metric for video generation},
  author={Unterthiner, Thomas and van Steenkiste, Sjoerd and Kurach, Karol and Marinier, Rapha{\"e}l and Michalski, Marcin and Gelly, Sylvain},
  year={2019}
}

@inproceedings{LPIPS ,
  title={The Unreasonable Effectiveness of Deep Features as a Perceptual Metric},
  author={Zhang, Richard and Isola, Phillip and Efros, Alexei A and Shechtman, Eli and Wang, Oliver},
  booktitle={CVPR},
  year={2018}
}

@inproceedings{deng2019arcface,
  title={Arcface: Additive angular margin loss for deep face recognition},
  author={Deng, Jiankang and Guo, Jia and Xue, Niannan and Zafeiriou, Stefanos},
  booktitle={Proceedings of the IEEE/CVF conference on computer vision and pattern recognition},
  pages={4690--4699},
  year={2019}
}

@inproceedings{chen2018lip,
  title={Lip movements generation at a glance},
  author={Chen, Lele and Li, Zhiheng and Maddox, Ross K and Duan, Zhiyao and Xu, Chenliang},
  booktitle={Proceedings of the European conference on computer vision (ECCV)},
  pages={520--535},
  year={2018}
}

@article{narvekar2011no,
  title={A no-reference image blur metric based on the cumulative probability of blur detection (CPBD)},
  author={Narvekar, Niranjan D and Karam, Lina J},
  journal={IEEE Transactions on Image Processing},
  volume={20},
  number={9},
  pages={2678--2683},
  year={2011},
  publisher={IEEE}
}

@InProceedings{huang2023vbench,
     title={{VBench}: Comprehensive Benchmark Suite for Video Generative Models},
     author={Huang, Ziqi and He, Yinan and Yu, Jiashuo and Zhang, Fan and Si, Chenyang and Jiang, Yuming and Zhang, Yuanhan and Wu, Tianxing and Jin, Qingyang and Chanpaisit, Nattapol and Wang, Yaohui and Chen, Xinyuan and Wang, Limin and Lin, Dahua and Qiao, Yu and Liu, Ziwei},
     booktitle={Proceedings of the IEEE/CVF Conference on Computer Vision and Pattern Recognition},
     year={2024}
 }

\end{document}